\documentclass[10pt,twocolumn,letterpaper]{article}

\usepackage{cvpr}
\usepackage{times}
\usepackage{epsfig}
\usepackage{graphicx}
\usepackage{amsmath}
\usepackage{amssymb}
\usepackage{color}
\usepackage{wrapfig}
\usepackage{authblk}

\usepackage{color,xcolor}
\usepackage{epsfig}
\usepackage{graphicx}

\usepackage{adjustbox}
\usepackage{array}
\usepackage{booktabs}
\usepackage{colortbl}
\usepackage{float,wrapfig}
\usepackage{hhline}
\usepackage{multirow}
\usepackage{subcaption} 
\usepackage[font={small}]{caption} 

\usepackage{amsmath,amsfonts,amsthm,amssymb}
\usepackage{bm}
\usepackage{nicefrac}
\usepackage{microtype}

\usepackage{changepage}
\usepackage{extramarks}
\usepackage{fancyhdr}
\usepackage{lastpage}
\usepackage{setspace}
\usepackage{soul}
\usepackage{xspace}

\usepackage{url}

\usepackage{algorithm, algorithmic}
\usepackage{enumerate}

\newcolumntype{L}[1]{>{\raggedright\let\newline\\\arraybackslash\hspace{0pt}}m{#1}}
\newcolumntype{C}[1]{>{\centering\let\newline\\\arraybackslash\hspace{0pt}}m{#1}}
\newcolumntype{R}[1]{>{\raggedleft\let\newline\\\arraybackslash\hspace{0pt}}m{#1}}

\newcommand{\fig}[1]{Fig.~\ref{#1}}
\newcommand{\tbl}[1]{Table~\ref{#1}}

\newcommand{\degree}{\ensuremath{^\circ}\xspace}
\newcommand{\ignore}[1]{}

\makeatletter
\DeclareRobustCommand\onedot{\futurelet\@let@token\@onedot}
\def\@onedot{\ifx\@let@token.\else.\null\fi\xspace}

\def\etal{\emph{et al}\onedot}

\makeatother

\definecolor{MyDarkBlue}{rgb}{0,0.08,1}
\definecolor{MyDarkGreen}{rgb}{0.02,0.6,0.02}
\definecolor{MyDarkRed}{rgb}{0.8,0.02,0.02}
\definecolor{MyDarkOrange}{rgb}{0.40,0.2,0.02}
\definecolor{MyPurple}{RGB}{111,0,255}
\definecolor{MyRed}{rgb}{1.0,0.0,0.0}
\definecolor{MyGold}{rgb}{0.75,0.6,0.12}
\definecolor{MyDarkgray}{rgb}{0.66, 0.66, 0.66}

\newcommand{\stress}[1]{\textbf{\textcolor{blue}{#1}}}

\newcommand{\jimei}[1]{}
\newcommand{\yasu}[1]{}
\newcommand{\duygu}[1]{}
\newcommand{\ersin}[1]{}

\newcommand{\orange}[1]{\textcolor{orange}{{#1}}}

\usepackage[pagebackref=true,breaklinks=true,letterpaper=true,colorlinks,bookmarks=false]{hyperref}

\cvprfinalcopy 

\def\cvprPaperID{1584} 

\ifcvprfinal\fi

\title{PlaneNet: Piece-wise Planar Reconstruction from a Single RGB Image}

\author{Chen Liu$^1$ \qquad Jimei Yang$^2$ \qquad Duygu Ceylan$^2$ \qquad Ersin Yumer$^3$ \qquad Yasutaka Furukawa$^4$ \\
$^1$Washington University in St. Louis \qquad $^2$Adobe Research \qquad $^3$Argo AI \qquad $^4$Simon Fraser University \\
{\tt\small chenliu@wustl.edu} \qquad {\tt\small \{jimyang,ceylan\}@adobe.com} \qquad {\tt\small meyumer@gmail.com} \qquad {\tt\small furukawa@sfu.ca}}

\begin{document}

\twocolumn[{
\maketitle
\vspace{-1.5em}
\centerline{
\includegraphics[width=\textwidth,trim={4pt 4pt 4pt 4pt}]{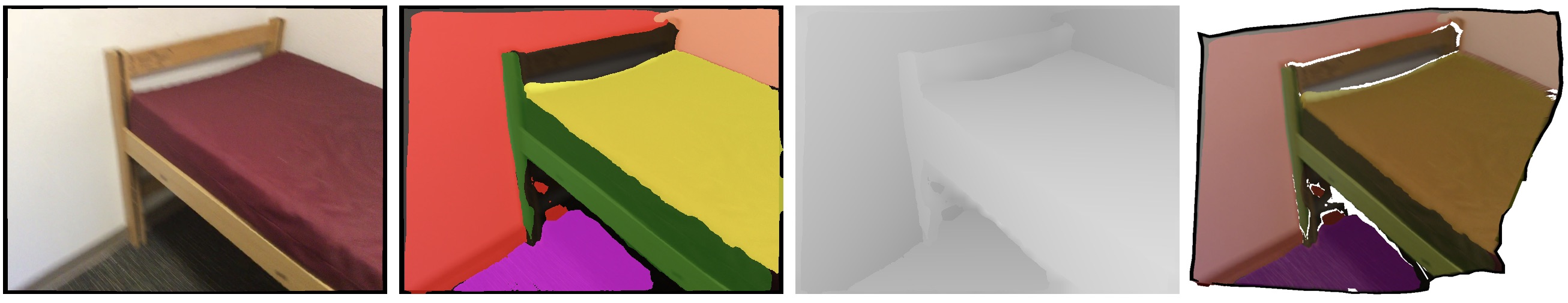}}
\captionof{figure}{
This paper proposes a deep neural architecture for piece-wise planar depthmap reconstruction from a single RGB image. From left to right, an input image, a piece-wise planar segmentation, a reconstructed depthmap, and a texture-mapped 3D model.
}
\label{fig:teaser}
\vspace{2em}
}]

\begin{abstract}
\footnotetext[1]{Work done during Chen Liu's internship at Adobe Research.}
This paper proposes a deep neural network (DNN) for piece-wise planar depthmap reconstruction from a single RGB image.
While DNNs have brought remarkable progress to single-image 
depth prediction, piece-wise planar depthmap reconstruction requires a structured geometry representation, and has been a difficult task to master even for DNNs.
The proposed end-to-end DNN learns to directly infer a set of plane parameters and corresponding plane segmentation masks from a single RGB image.
%
We have generated more than 50,000 piece-wise planar depthmaps for training and testing from ScanNet, a large-scale RGBD video database.
Our qualitative and quantitative evaluations demonstrate that the proposed approach outperforms baseline methods in terms of both plane segmentation and depth estimation accuracy. 
%
To the best of our knowledge, this paper presents the first end-to-end neural architecture for piece-wise planar reconstruction from a single RGB image.
Code and data are available at \url{https://github.com/art-programmer/PlaneNet}.
\end{abstract}
\vspace{-15pt}
\section{Introduction}
Human vision has a remarkable perceptual capability in understanding high-level scene structures. Observing a typical indoor scene (e.g.,  Fig.~\ref{fig:teaser}), we can instantly parse a room into a few number of dominant planes (e.g., a floor, walls, and a ceiling), perceive major surfaces for a furniture, or recognize a horizontal surface at the table-top.
Piece-wise planar geometry understanding would be a key for many applications in emerging domains such as robotics or augmented reality (AR). For instance, a robot needs to identify the extent of a floor to plan a movement, or a table-top segmentation to place objects.
In AR applications, planar surface detection is becoming a fundamental building block for placing virtual objects on a desk~\cite{Karsch2014tog}, replacing floor textures, or hanging artworks on walls for interior remodeling.
%
A fundamental problem in Computer Vision is to develop a computational algorithm that masters similar perceptual capability to enable such applications.

With the surge of deep neural networks, single image depthmap inference~\cite{eigen2014depth,eigen2015predicting,laina2016deeper,wang2016surge,xu2017multi} and room layout estimation~\cite{lee2017roomnet} have been active areas of research. However, to our surprise, little attention has been given to the study of \emph{piece-wise planar depthmap reconstruction}, mimicking this remarkable human perception in a general form. 
%
%
The main challenge is that the piece-wise planar depthmap requires structured geometry representation (i.e., a set of plane parameters and their segmentation masks). In particular, we do not know the number of planes to be inferred, and the order of planes to be regressed in the output feature vector, making the task challenging even for deep neural networks.

This paper proposes a novel deep neural architecture ``PlaneNet'' that learns to directly produce a set of plane parameters and probabilistic plane segmentation masks from a single RGB image.
%
Following a recent work on point-set generation~\cite{fan2016point}, we define a loss function that is agnostic to the order of planes. We further control the number of planes by allowing probabilistic plane segmentation masks to be all 0~\cite{tulsiani2016learning}.
The network also predicts a depthmap at non-planar surfaces, whose loss is defined through the probabilistic segmentation masks to allow back-propagation.
%
We have generated more than 50,000 piece-wise planar depthmaps from ScanNet~\cite{dai2017scannet} as ground-truth 
by fitting planes to 3D points and projecting them to images.
Qualitative and quantitative evaluations show that
our algorithm produces significantly better plane segmentation results than
the current state-of-the-art.
Furthermore, our depth prediction accuracy is on-par or even superior to the existing single image depth inference techniques that are specifically trained for this task.
\section{Related work}\label{section:related_work}

\noindent
\textbf{Multi-view piece-wise planar reconstruction.}
Piece-wise planar depthmap reconstruction was once an active research topic in multi-view 3D reconstruction~\cite{furukawa2009manhattan,sinha2009piecewise,gallup2010piecewise,zebedin2008fusion}. The task is to infer a set of plane parameters and assign a plane-ID to each pixel. Most existing methods first reconstruct precise 3D points, perform plane-fitting to generate plane hypotheses, then solve a global inference problem to reconstruct a piece-wise planar depthmap. 
Our approach learns to directly infer plane parameters and plane segmentations from a single RGB image.

\vspace{0.1cm}
\noindent
\textbf{Learning based depth reconstruction.} Saxena \etal~\cite{saxena2006learning} pioneered a learning based approach for depthmap inference from a single image. 
With the surge of deep neural networks, numerous CNN based approaches have been proposed~\cite{eigen2014depth,li2015depth,roy2016monocular}. However, most techniques simply produce an array of depth values (i.e., depthmap) without plane detection or segmentation.
More recently, Wang \etal~\cite{wang2016surge} enforce planarity in depth (and surface normal) predictions by inferring pixels on planar surfaces.
This is the closest work to ours. However, they only produce a binary segmentation mask (i.e., if a pixel is on a planar surface or not) without plane parameters or instance-level plane segmentation.




\vspace{0.1cm}
\noindent
\textbf{Layout estimation.} Room layout estimation also aims at predicting dominant planes in a scene (e.g., walls, floor, and ceiling). Most traditional approaches~\cite{hedau2009recovering,lee2009geometric,gupta2010estimating,schwing2012efficient,fouhey2014people,wang2013discriminative} rely on image processing heuristics to estimate vanishing points of a scene, and aggregate low-level features by a global optimization procedure. Besides low-level features, high-level information has been utilized, such as human poses~\cite{chao2013layout,fouhey2014people} or semantics~\cite{chao2013layout,bao2014understanding}. Attempts have been made to go beyond room structure, and predict object geometry~\cite{gupta2010estimating,wang2013discriminative,bao2014understanding,zhang2014panocontext}. However, the reliance on hand-crafted features makes those methods less robust, and the Manhattan World assumption limits their operating ranges. Recently, Lee \etal~\cite{lee2017roomnet} proposed an end-to-end deep neural network, RoomNet, which simultaneously classifies a room layout type and predicts corner locations.
However, their framework is not applicable to general piece-wise planar scenes.

\vspace{0.1cm}
\noindent \textbf{Line analysis.} 
Single image 3D reconstruction of line drawings date back to the 60s. The earliest attempt is probably the Robert's system~\cite{roberts1963machine}, which inspired many follow-up works~\cite{sugihara1986machine,xue2012example}. In real images, extraction of line drawings is challenging. Statistical analysis of line directions, junctions, or image segments have been used to enable 3D reconstruction for architectural scenes~\cite{ramalingam2013lifting} or indoor panoramas~\cite{yang2016efficient}. Attributed grammar was used to parse an image into a hierarchical graph for 3D reconstruction~\cite{liu2014single}. However, these approaches require hand-crafted features, grammar specification, or algorithmic rules. Our approach is purely data-driven harnessing the power of deep neural networks.

\section{PlaneNet}
\label{sect:model}
\begin{figure*}
    \centering
    \includegraphics[width=\textwidth]{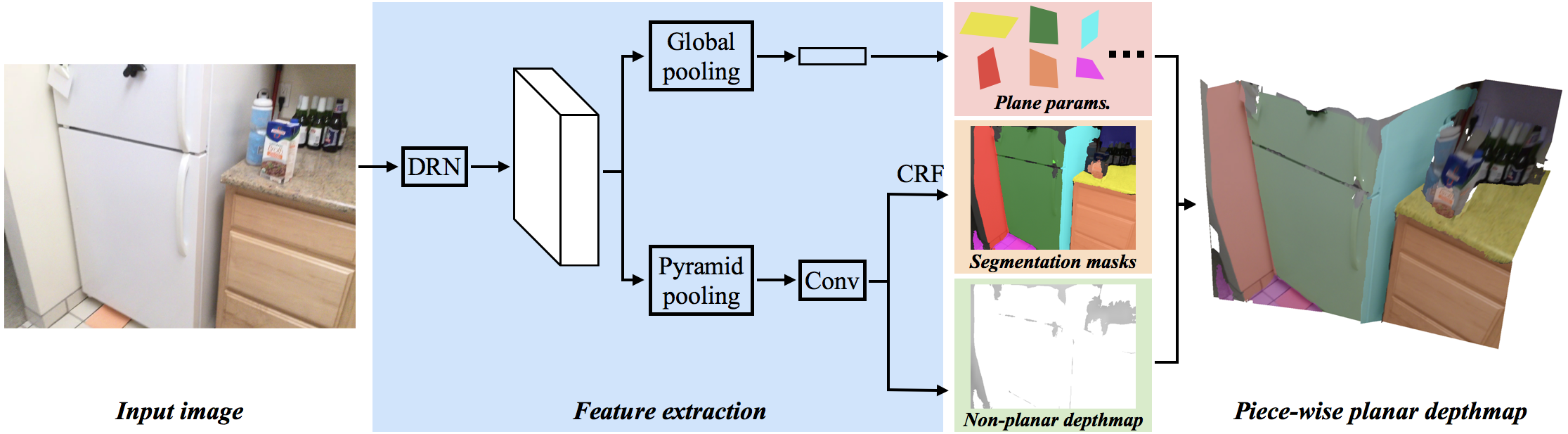}
\caption{PlaneNet predicts plane parameters, their probabilistic segmentation masks, and a non-planar depthmap from a single RGB image.
}
    \label{fig:pipeline}
    \vspace{-5pt}
\end{figure*}

We build our network upon Dilated Residual Networks (DRNs)~\cite{yu2017dilated, chen2016deeplab} (See \fig{fig:pipeline}), which is a flexible framework for both global tasks (e.g., image classification) and pixel-wise prediction tasks (e.g., semantic segmentation).
Given the high-resolution final feature maps from DRN, we compose three output branches for the three prediction tasks. 

\vspace{0.1cm}
\noindent \textbf{Plane parameters:} For each scene, we predict a fixed number ($K$) of planar surfaces $\mathcal{S}=\{S_1, \cdots S_K\}$. Each surface $S_i$ is specified by the three plane parameters $P_i$ (i.e., encoding a normal and an offset). We use $D_i$ to denote a depth image, which can be inferred from the parameters $P_i$~\footnote{The depth value calculation requires camera intrinsic parameters, which can be estimated via vanishing point analysis, for example. In our experiments, intrinsics are given for each image through the database information.}.

\vspace{0.1cm}
\noindent \textbf{Non-planar depthmap:} We model non-planar structures and infer its geometry as a standard depthmap. With abuse of notation, we treat it as the $(\textit{K+1})^{th}$ surface and denote the depthmap as $D_{K+1}$. This does not explain planar surfaces.

\vspace{0.1cm}
\noindent \textbf{Segmentation masks:} The last output is the probabilistic segmentation masks for the $K$ planes ($M_1, \cdots M_K$) and the non-planar depthmap ($M_{K+1}$).

\vspace{0.2cm}
\noindent To summarize, the network predicts 1) plane parameters ($P_1, \cdots, P_K$), 2) a non-planar depthmap ($D_{K+1}$), and 3) probabilistic segmentation masks ($M_1, \cdots, M_{K+1})$. We now explain more details and the loss function for each task.

%

\subsection{Plane parameter branch}
The plane parameter branch starts with a global average-pooling to reduce the feature map size to 1x1~\cite{yu2017dilated}, followed by a fully connected layer to produce $K\times3$ plane parameters. We do not know the number of planes as well as their order in this prediction task.
By following prior works~\cite{fan2016point,tulsiani2016learning}, we
predict a constant number (K) of planes, then allow some predictions to be invalid by letting the corresponding probabilistic segmentation masks to be 0.
Our ground-truth generation process (See Sect.~\ref{datasets}) produces at most 10 planes for most examples, thus we set $K=10$ in our experiments.
We define an order-agnostic loss function based on the Chamfer distance metric for the regressed plane parameters:
\begin{equation}
\mathcal{L}^P = \sum_{i=1}^{K^*}\min_{j \in [1, K]}\left\|P_i^* - P_j\right\|^2_2.
\end{equation}
The parameterization $P_i$ is given by the 3D coordinate of the point that is closest to the camera center on the plane. 
$P_i^*$ is the ground-truth. $K^*$ is the number of ground-truth planes. 

\subsection{Plane segmentation branch}
The branch
starts with a pyramid pooling module~\cite{zhao2016pyramid}, followed by a convolutional layer to produce $K+1$ channel likelihood maps for planar and non-planar surfaces. We append a dense conditional random field (DCRF) module based on the fast inference algorithm proposed by Krahenbuhl and Koltun~\cite{krahenbuhl2011efficient}, and jointly train the DCRF module with the precedent layers as in Zheng \etal~\cite{zheng2015conditional}. We set the number of meanfield iterations to 5 during training and to 10 during testing. Bandwidths of bilateral filters are fixed for simplicity.
%
We use a standard softmax cross entropy loss to supervise the segmentation training:
\begin{equation}
    \mathcal{L}^M = \sum_{i=1}^{K+1}\sum_{p \in I}(\textbf{1}(M^{*(p)}=i)\log(1 - M_i^{(p)})) 
\end{equation}
The internal summation is over the image pixels ($I$), where $M^{(p)}_{i}$ denotes the probability of pixel $p$ belonging to the $i^{th}$ plane. $M^{*(p)}$ is the ground-truth plane-id for the pixel.

\subsection{Non-planar depth branch}
%
The
branch shares the same pyramid pooling module, followed by a convolution layer to produce a $1$-channel depthmap. Instead of defining a loss specifically for non-planar regions, we found that
exploiting the entire ground-truth depthmap makes the overall training more effective. Specifically, 
we define the loss as the sum of squared depth differences between the ground-truth and either a predicted plane or a non-planar depthmap, weighted by probabilities:
%
\begin{equation}
    \mathcal{L}^D = \sum_{i=1}^{K+1}\sum_{p \in I}(M_i^{(p)} (D_i^{(p)} - D^{*(p)})^2)
    \label{equ:depth-loss}
\end{equation}
$D_i^{(p)}$ denotes the depth value at pixel $p$,
while $D^{*(p)}$ is the ground truth depth value.
%





\section{Datasets and implemenation details}
\label{datasets}

We have generated 51,000 ground-truth piece-wise planar depthmaps (50,000 training and 1,000 testing) from ScanNet~\cite{dai2017scannet}, a large-scale indoor RGB-D video database. A depthmap in a single RGB-D frame contains holes and the quality deteriorates at far distances. Our approach for ground-truth generation is to directly fit planes to a consolidated mesh and project them back to individual frames, while also exploiting the associated semantic annotations.

Specifically, for each sub mesh-models of the same semantic label, we treat mesh-vertices as points and repeat extracting planes by RANSAC with replacement. The inlier distance threshold is $5cm$, and the process continues until
$90\%$ of the points are covered.
We merge two (not necessarily adjacent) planes that span different semantic labels if the plane normal difference is below $20\degree$, and if the larger plane fits the smaller one with the mean distance error below $5cm$.
We project each triangle to individual frames if the three vertices are fitted by the same plane. After projecting all the triangles, we keep only the planes whose projected area is larger than $1\%$ of an image. We discard entire frames if the ratio of pixels covered by the planes is below 50\%.
For training samples, we randomly choose 90\% of the scenes from ScanNet, subsample every 10 frames, compute piece-wise planar depthmaps with the above procedure, then use the final random sampling to produce 50,000 examples. The same procedure generates 1,000 testing examples from the remaining 10\% of the scenes.


We have implemented PlaneNet using TensorFlow~\cite{abadi2016tensorflow} based on DeepLab~\cite{chen2016deeplab}. Our system is a 101-layer ResNet~\cite{He2016CVPR} with Dilated Convolution, while we have followed a prior work and modified the first few layers to deal with the degridding issue~\cite{yu2017dilated}. The final feature map of the DRN contains 2096 channels.
We use the Adam optimizer~\cite{kingma2014adam} with the initial learning rate set to 0.0003. The input image, the output plane segmentation masks, and the non-planar depthmap have a resolution of 256x192. We train our network for 50 epochs on the 50,000 training samples.



\section{Experimental results}
\begin{figure*}[p]
    \centering
    \includegraphics[width=0.92\linewidth]{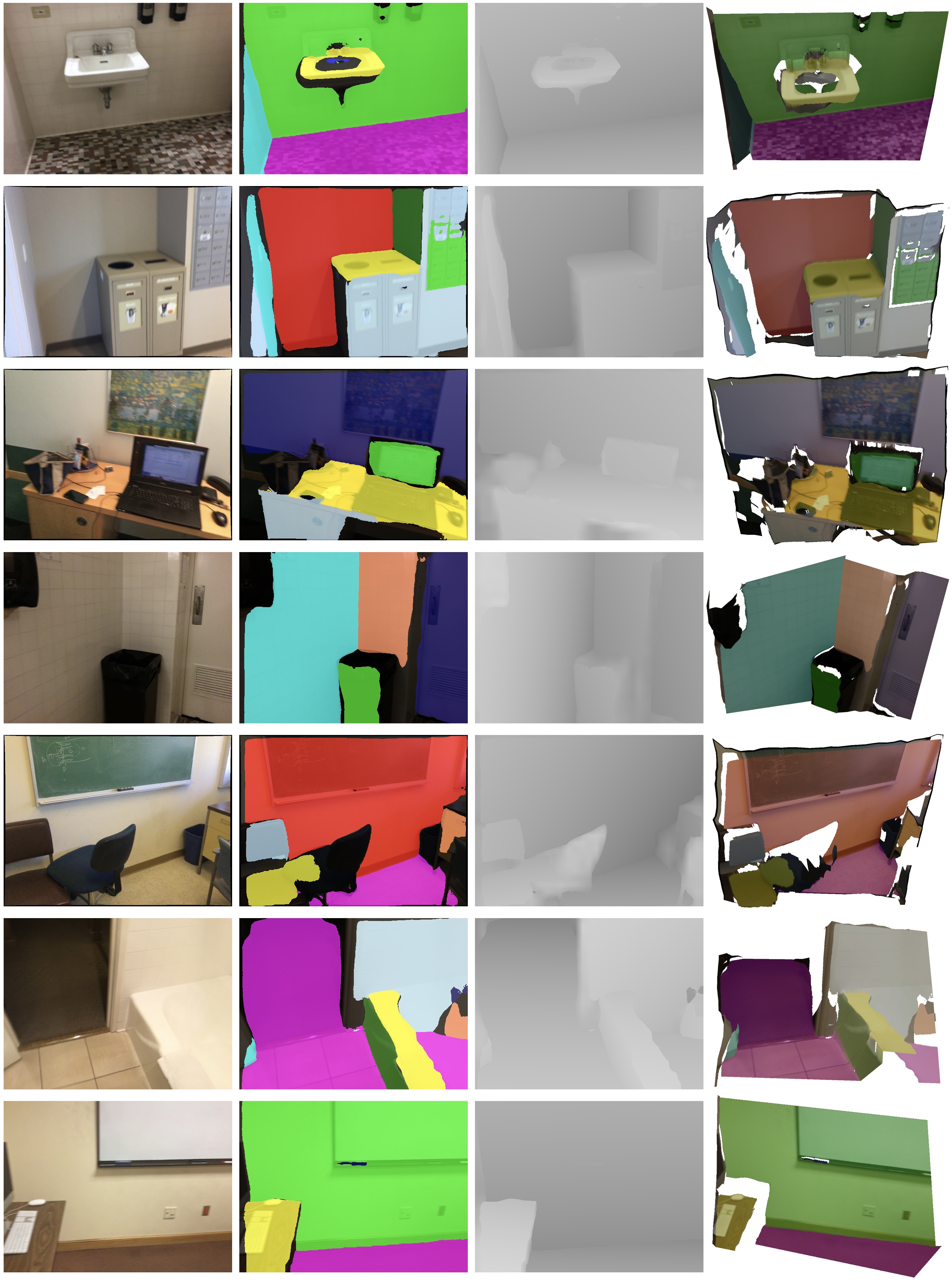}
\caption{Piece-wise planar depthmap reconstruction results by PlaneNet. From left to right: input image, plane segmentation, depthmap reconstruction, and 3D rendering of our depthmap. In the plane segmentation results, the black color shows non-planar surface regions.
}
    \label{fig:results}
    \vspace{-5pt}
\end{figure*}

Figure~\ref{fig:results} shows our reconstruction results for a variety of scenes. Our end-to-end learning framework has successfully recovered piece-wise planar and semantically meaningful structures, such as floors, walls, table-tops, or a computer screen, from a single RGB image.
We include many more examples in the supplementary material. 
We now provide quantitative evaluations on the plane segmentation accuracy and the depth reconstruction accuracy against the competing baselines, followed by more analyses of our results.


\subsection{Plane segmentation accuracy}
\begin{figure*}[tb]
    \centering
    \includegraphics[width=0.9\textwidth]{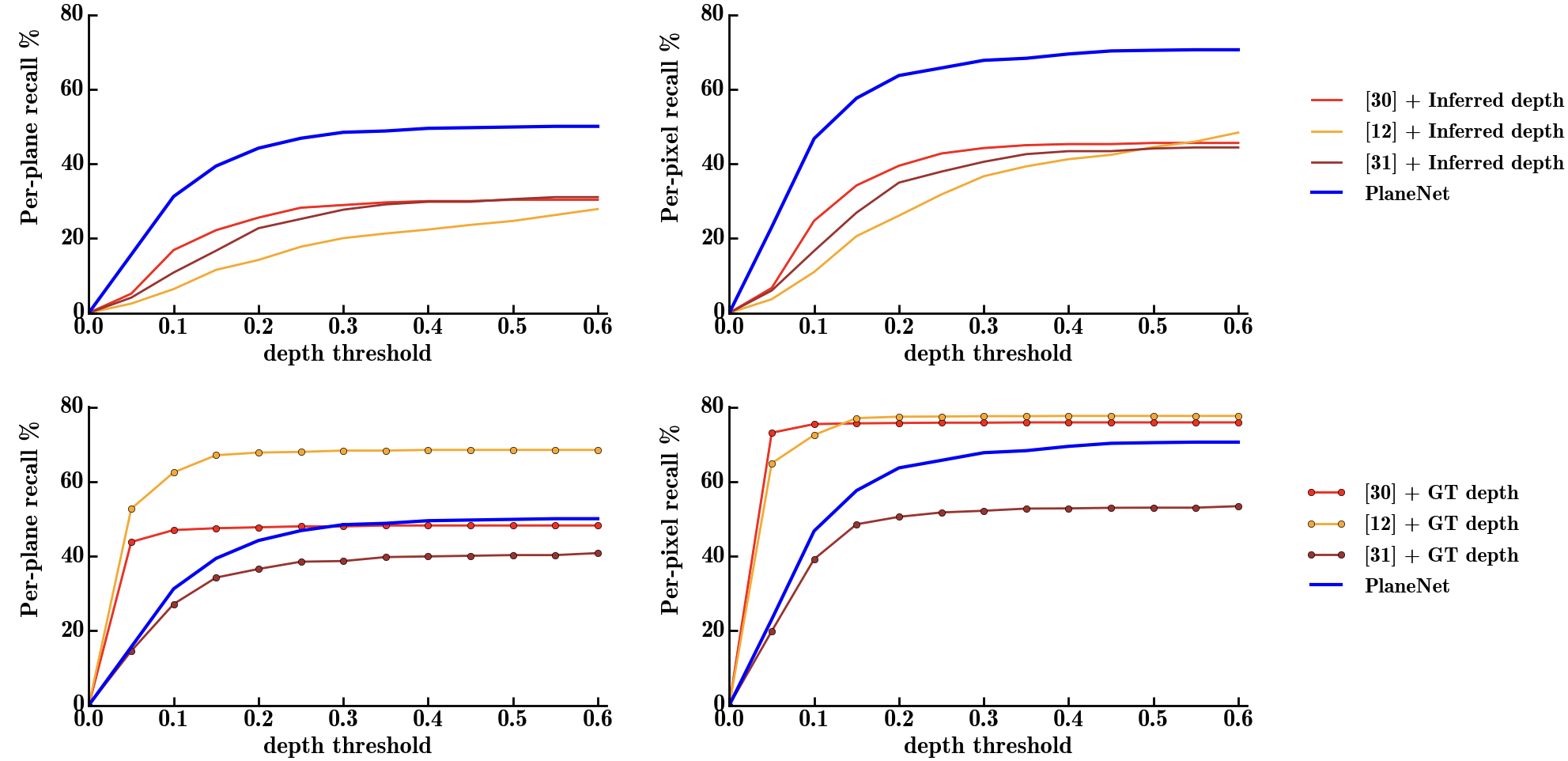}
    
\caption{Plane segmentation accuracy against competing baselines that use 3D points as input~\cite{silberman2012indoor,furukawa2009manhattan,sinha2009piecewise}. Either ground-truth depthmaps or inferred depthmaps (by a DNN-based system) are used as their inputs.
PlaneNet outperforms all the other methods that use inferred depthmaps. Surprisingly, PlaneNet is even better than many other methods that use ground-truth depthmaps.
}
    \label{fig:piece-wise}
    \vspace{-5pt}
\end{figure*}
Piece-wise planar reconstruction from a single RGB image is a challenging problem. While existing approaches have produced encouraging results~\cite{fouhey2014unfolding,liu2014single,ramalingam2013lifting},
they are based on hand-crafted features and algorithmic designs, and may not match against big-data and deep neural network (DNN) based systems.
Much better baselines would then be piece-wise planar depthmap reconstruction techniques from 3D points~\cite{furukawa2009manhattan,sinha2009piecewise,gallup2010piecewise,zebedin2008fusion}, where input 3D points are either given by the ground-truth depthmaps or inferred by a state-of-the-art DNN-based system~\cite{laina2016deeper}.

In particular, to infer depthmaps, we have used a variant of PlaneNet which only has the pixel-wise depthmap branch, while following Eigen \etal~\cite{eigen2015predicting} to change the loss. \tbl{tbl:comparison} shows that this network, PlaneNet (Depth rep.), outperforms the current top-performers on the NYU benchmark~\cite{silberman2012indoor}.

For piece-wise planar depthmap reconstruction, we have used the following three baselines from the literature.

\vspace{0.1cm}
\noindent $\bullet$ ``NYU-Toolbox" is a plane extraction algorithm from the official NYU toolbox~\cite{silberman2012indoor} that extracts plane hypotheses using RANSAC, and optimizes the plane segmentation via a Markov Random Field (MRF) optimization.

\vspace{0.1cm}
\noindent $\bullet$ Manhattan World Stereo (MWS)~\cite{furukawa2009manhattan} is very similar to NYU-Toolbox except that MWS employs the Manhattan World assumption in extracting planes and exploits vanishing lines in the pairwise terms to improve results.

\vspace{0.1cm}
\noindent $\bullet$ Piecewise Planar Stereo (PPS)~\cite{sinha2009piecewise} relaxes the Manhattan World assumption of MWS, and uses vanishing lines to generate better plane proposals. Please see the supplementary document for more algorithmic details on the baselines.



\vspace{0.1cm}
%
%
%
%
Figure~\ref{fig:piece-wise} shows the evaluation results 
%
on two recall metrics. The first metric is the percentage of correctly predicted ground-truth planes. We consider a ground-truth plane being correctly predicted, if one of the inferred planes has 1) more than 0.5 Intersection over Union (IOU) score and 2) the mean depth difference over the overlapping region is less than a threshold. We vary this threshold from 0 to 0.6m with an increment of 0.05m to plot graphs. The second recall metric is simply the percentage of pixels that are in such overlapping regions where planes are correctly predicted.
%
%
The figure shows that PlaneNet is significantly better than all the competing methods when inferred depthmaps are used.
PlaneNet is even better than some competing methods that use ground-truth depthmaps. This demonstrates the effectiveness of our approach, learning to infer piece-wise planar structures from many examples.

\begin{figure*}
\centering
    \includegraphics[width=\linewidth]{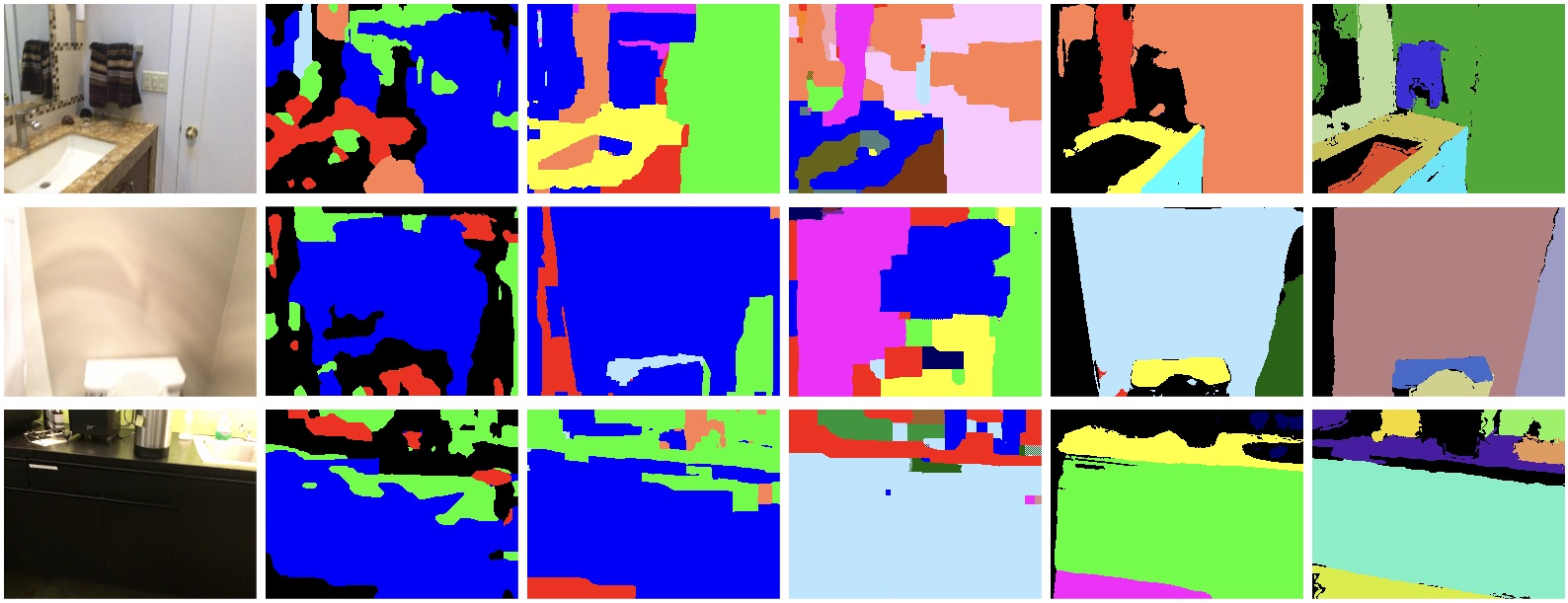}
\caption{Qualitative comparisons between PlaneNet and existing methods that use inferred depthmaps as the inputs. From left to right: an input image, plane segmentation results for ~\cite{silberman2012indoor}, ~\cite{furukawa2009manhattan}, ~\cite{sinha2009piecewise}, and PlaneNet, respectively, and the ground-truth.}
    \label{fig:comparison}
    \vspace{-5pt}
\end{figure*}

Figure~\ref{fig:comparison} shows qualitative comparisons against existing methods with inferred depthmaps. PlaneNet produces significantly better plane segmentation results, while 
existing methods often generate many redundant planes where depthmaps are noisy, and fail to capture precise boundaries where the intensity edges are weak.

%
 


\subsection{Depth reconstruction accuracy}
While the capability to infer a plane segmentation mask and precise plane parameters is the key contribution of the work, it is also interesting to compare against depth prediction methods. This is to ensure that our structured depth prediction does not compromise per-pixel depth prediction accuracy.
PlaneNet makes (K+1) depth value predictions at each pixel. We pick the depth value with the maximum probability in the segmentation mask to define our depthmap.
%

Depth accuracies are evaluated on the NYUv2 dataset at 1) planar regions, 2) boundary regions, and 3) the entire image, against three competing baselines.
~\footnote{Following existing works, we choose NYUv2 to evaluate depth accuracy and consider only the valid 561x427 area as the entire image evaluation.}
%
Eigen-VGG~\cite{eigen2015predicting} is a convolutional architecture to predict both depths and surface normals. SURGE~\cite{wang2016surge} is a more recent depth inference network that optimizes planarity. FCRN is the current state-of-the-art single-image depth inference network~\cite{laina2016deeper}~\footnote{The numbers are different from the numbers reported in~\cite{laina2016deeper} since ~\cite{laina2016deeper} evaluate on the original resolution, 640x480, and their numbers are influenced by the issue reported at https://github.com/iro-cp/FCRN-DepthPrediction/issues/42.}. 

Depthmaps in NYUv2 are very noisy and ground-truth plane extraction does not work well. Thus, we fine-tune our network using only the depth loss (\ref{equ:depth-loss}). Note that the key factor in this training is that the network is trained to generate a depthmap through our piece-wise planar depthmap representation.
To further verify the effects of this representation, we have also fine-tuned our network in the standard per-pixel depthmap representation by disabling the plane parameter and the plane segmentation branches. In this version, denoted as ``PlaneNet (Depth rep.)'',
the entire depthmap is predicted in the $(K+1)^{th}$ depthmap ($D_{K+1}$).

Table~\ref{tbl:comparison} shows the depth prediction accuracy on various metrics introduced in the prior work~\cite{eigen2014depth}.
The left five metrics provide different error statistics such as relative difference (Rel) or rooted-mean-square-error (RMSE) on the average per-pixel depth errors.
The right three metrics provide the ratio of pixels, for which the relative difference between the predicted and the ground-truth depths is below a threshold.
%
The table demonstrates that PlaneNet outperforms the state-of-the-art of single-image depth inference techniques.
%
As observed in prior works~\cite{wang2016surge,chakrabarti2016depth}, the planarity constraint makes differences in the depth prediction task, and the improvements are more significant when our piece-wise planar representation is enforced by our network.




\begin{table*}[t]
\caption{Depth accuracy comparisons over the NYUv2 dataset.}
\label{tbl:comparison}
  \centering
  \begin{tabular}{l|ccccc|ccc}
    \toprule
  & \multicolumn{5}{c}{Lower the better (LTB)} & \multicolumn{3}{c}{Higher the better (HTB)} \\
  Method & Rel & Rel(sqr) & $log_{10}$ & $RMSE_{iin}$ & $RMSE_{log}$ & 1.25 & $1.25^2$ & $1.25^3$\\
  \midrule
  \multicolumn{9}{c}{Evaluation over planar regions} \\
  \midrule
Eigen-VGG~\cite{eigen2015predicting} & 0.143 & 0.088 & 0.061 & 0.490 & 0.193 & 80.1 & 96.4 & \stress{99.3}\\
  SURGE~\cite{wang2016surge} & 0.142 & 0.087 & 0.061 & 0.487 & 0.192 & 80.2 & 96.6 & \stress{99.3}\\
  FCRN~\cite{laina2016deeper} & 0.140 & 0.087 & 0.065 & 0.460 & 0.183 & 79.2 & 95.6 & 99.0\\
\midrule
PlaneNet (Depth rep.) & \orange{0.130} & \orange{0.080} & \stress{0.054} & \orange{0.399} & \orange{0.156} & \stress{84.4} & \orange{96.7} & 99.2 \\
PlaneNet & \stress{0.129} & \stress{0.079} & \stress{0.054} & \stress{0.397} & \stress{0.155} & \orange{84.2} & \stress{96.8} & 99.2 \\
\midrule
\specialrule{2pt}{2pt}{1pt}
  \multicolumn{9}{c}{Evaluation over edge areas} \\
  \midrule
Eigen-VGG~\cite{eigen2015predicting} & 0.165 & 0.137 & 0.073 & 0.727 & 0.228 & 72.9 & 74.3 & 98.7\\
SURGE~\cite{wang2016surge} & 0.162 & 0.133 & 0.071 & 0.697 & 0.221 &  74.7 & 94.7 & 98.7 \\ 
FCRN~\cite{laina2016deeper} & 0.154 & 0.111 & 0.073 & 0.548 & 0.208 & 74.7 & 94.1 & 98.5 \\
\midrule
PlaneNet (Depth rep.) & \stress{0.145} & \stress{0.099} & \stress{0.061} & \orange{0.480} & \orange{0.179} & \stress{80.9} & \orange{95.9} & \orange{99.0} \\
PlaneNet & \stress{0.145} & \stress{0.099} & \stress{0.061} & \stress{0.479} & \stress{0.178} & \orange{80.7} & \stress{96.1} & \stress{99.1} \\
\midrule
\specialrule{2pt}{2pt}{1pt}
  \multicolumn{9}{c}{Evaluation over the entire image} \\
  \midrule
Eigen-VGG~\cite{eigen2015predicting} & 0.158 & 0.121 & 0.067 & 0.639 & 0.215 & 77.1 & 95.0 & 98.8\\
SURGE~\cite{wang2016surge} & 0.156 & 0.118 & 0.067 & 0.643 & 0.214 & 76.8 & 95.1 & \stress{98.9} \\ 
FCRN~\cite{laina2016deeper} & 0.152 & 0.119 & 0.072 & 0.581 & 0.207 & 75.6 & 93.9 & 98.4 \\
\midrule
PlaneNet (Depth rep.) & \orange{0.143} & \stress{0.107} & \stress{0.060} & \orange{0.518} & \orange{0.180} & \stress{81.3} & \orange{95.5} & 98.7 \\
PlaneNet & \stress{0.142} & \stress{0.107} & \stress{0.060} & \stress{0.514} & \stress{0.179} & \orange{81.2} & \stress{95.7} & \stress{98.9} \\
  \bottomrule
  \end{tabular}
\vspace{-5pt}
\end{table*}


\subsection{Plane ordering consistency}
\label{sect:ordering}
The ordering ambiguity is a challenge for piece-wise depthmap inference. We found that PlaneNet automatically learns a consistent ordering without supervision, for example, the floor is always regressed as the second plane. In Fig.~\ref{fig:results}, colors in the plane segmentation results are defined by the order of the planes in the network output. Although the ordering loses consistency for small objects or extreme camera angles, major common surfaces such as the floor and walls have a consistent ordering in most cases.

\begin{figure}
    \centering
    \includegraphics[width=\linewidth]{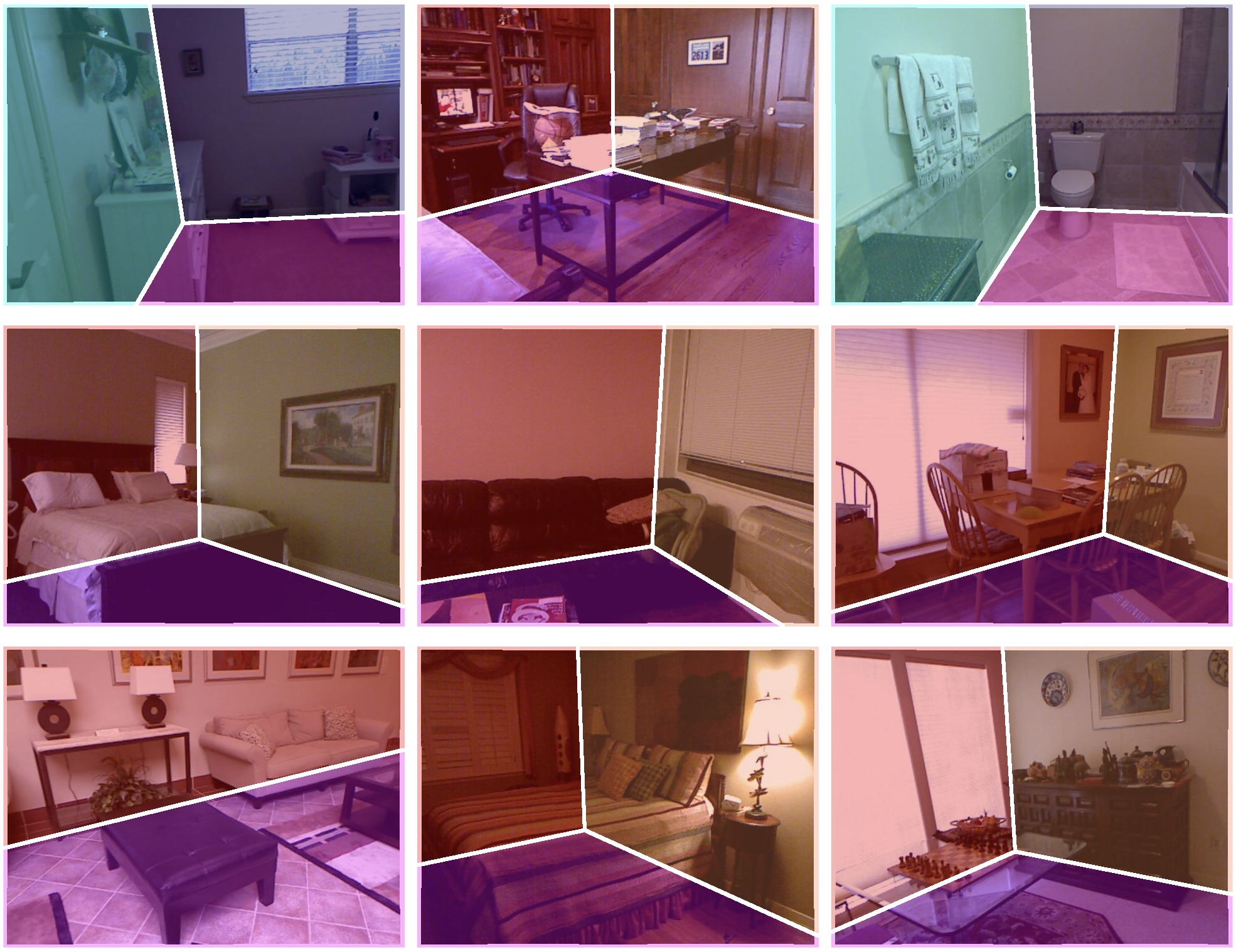}
\caption{Room layout estimations. We have exploited the ordering consistency in the predicted planes to infer room layouts. 
}
    \label{fig:layout}
    \vspace{-5pt}
\end{figure}

We have exploited this property and implemented a simple room layout estimation algorithm.
%
More specifically, we look at reconstruction examples and manually select the entries of planes that correspond to the ceiling, the floor, and the left/middle/right walls. 
For each possible room layout configuration~\cite{lee2017roomnet}, (e.g., a configuration with a floor, a left wall, and a middle wall visible),
we construct a 3D concave hull based on the plane parameters and project it back to the image to generate a room-layout. We measure the score of the configuration by the number of pixels, where the constructed room layout and the inferred plane segmentation (determined by the winner-takes-all) agree.
We pick the constructed room layout with the best score as our prediction.
%
Figure~\ref{fig:layout} shows that our algorithm is able to generate reasonable room layout estimations even when the scene is cluttered and contain many occluding objects.
%
Table~\ref{tbl:layout} shows the quantitative evaluations on the NYUv2 303 dataset~\cite{zhang2013estimating}, where our method is comparable to existing techniques which are designed specifically for this task.~\footnote{RoomNet paper~\cite{lee2017roomnet} does not provide code or evaluation numbers for the NYUv2 benchmark.
We have implemented their system using Torch7 and trained on LSUN dataset as described in their paper.}
%

\begin{table}
\caption{Room layout estimations. Quantitative evaluations against the top-performers over the NYUv2 303 dataset. 
}
\vspace{-5pt}
\label{tbl:layout}
  \centering
  \begin{tabular}{|c||c|c|}
  \hline
  & Input & Layout error \\
  \hline
  Schwing \etal~\cite{schwing2012efficient} & RGB & $13.66\%$ \\
  \hline
  Zhang \etal~\cite{zhang2013estimating} & RGB & $13.94\%$ \\
  \hline
  Zhang \etal~\cite{zhang2013estimating} & RGB+D & $8.04\%$ \\
  \hline
  RoomNet~\cite{lee2017roomnet} & RGB & $12.96\%$ \\
  \hline
  PlaneNet & RGB & $12.64\%$ \\
  \hline
  \end{tabular}
  \vspace{-5pt}
\end{table}

\subsection{Failure modes}
\begin{figure}
    \centering
    \includegraphics[width=0.9\linewidth]{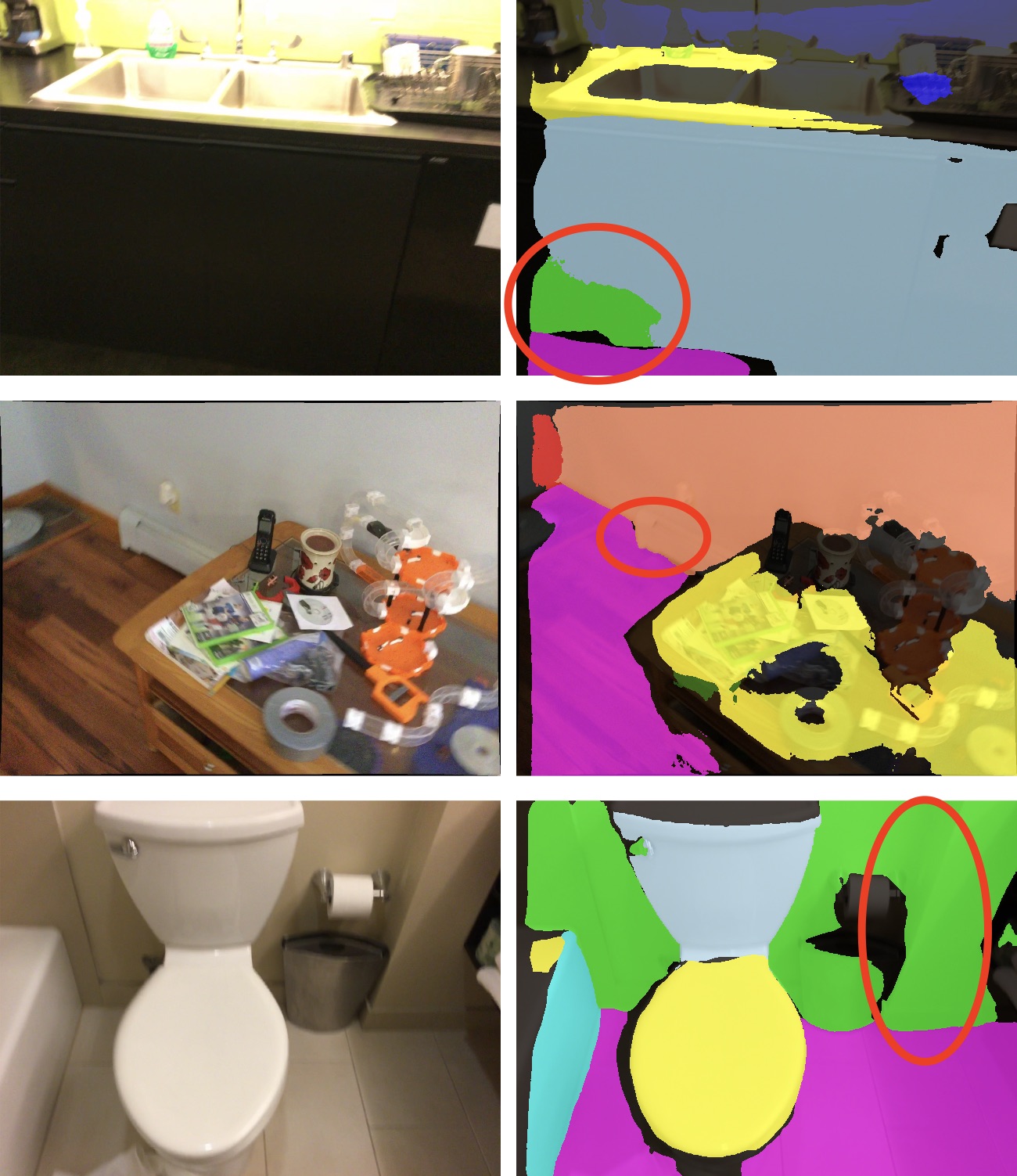}
\caption{Typical failure modes occur in the absence of enough image texture cues or at the presence of small objects and clutter.}
    \label{fig:failures}
    \vspace{-5pt}
\end{figure}

While achieving promising results on most images, PlaneNet has some failure modes as shown in \fig{fig:failures}. In the first example, PlaneNet generates two nearly co-planar vertical surfaces in the low-light region below the sink. In the second example, it cannot distinguish a white object on the floor from a white wall. In the third example, it misses a column structure on a wall due to the presence of object clutter.
While the capability to infer precise plane parameters is already super-human, 
there is a lot of room for improvement on
the planar segmentation, especially in the absence of texture information or at the presence of clutter.
%

\section{Applications}
\begin{figure}
    \centering
    \includegraphics[width=0.9\linewidth]{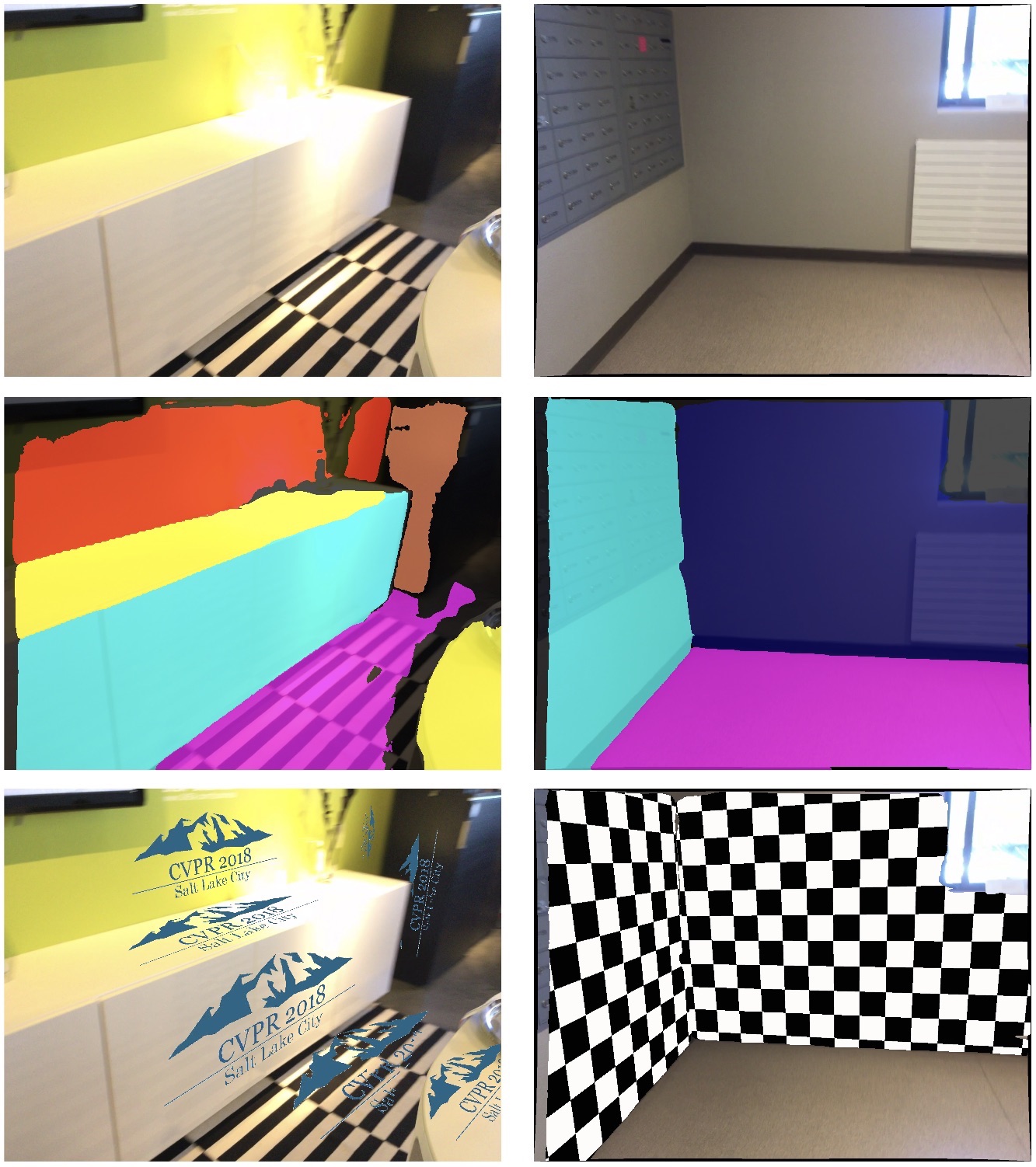}
\caption{Texture editing applications. From top to bottom, an input image, a plane segmentation result, and an edited image.
}
    \label{fig:applications}
    \vspace{-5pt}
\end{figure}
Structured geometry reconstruction is important for many application in Augmented Reality.
We demonstrate two image editing applications enabled by our piece-wise planar representation: texture insertion and replacement (see Fig.~\ref{fig:applications}).
%
We first extract Manhattan directions by using the predicted plane normals through a standard voting scheme~\cite{furukawa2009manhattan}. Given a piece-wise planar region, we define an axis of its UV coordinate by the Manhattan direction that is the most parallel to the plane, while the other axis is simply the cross product of the first axis and the plane normal.
Given a UV coordinate, we insert a new texture by alpha-blending or completely replace a texture with a new one. Please see the supplementary material and the video for more AR application examples.
\section{Conclusion and future work}
This paper proposes PlaneNet, the first deep neural architecture for piece-wise planar depthmap reconstruction from a single RGB image. PlaneNet learns to directly infer a set of plane parameters and their probabilistic segmentation masks. The proposed approach significantly outperforms competing baselines in the plane segmentation task. It also advances the state-of-the-art in the single image depth prediction task.
An interesting future direction is to go beyond the depthmap framework and tackle structured geometry prediction problems in a full 3D space.
\section{Acknowledgement}

This research is partially supported by National Science Foundation under grant IIS 1540012 and IIS 1618685, Google Faculty Research Award, and Adobe gift fund. We thank Nvidia for a generous GPU donation.

\clearpage
\newpage
{\small
\bibliographystyle{ieee}
\bibliography{main}
}

\end{document}